\documentclass[pdflatex,sn-mathphys]{sn-jnl}

\jyear{2022}%

\theoremstyle{thmstyleone}%
\theoremstyle{thmstyletwo}%

\theoremstyle{thmstylethree}%

\usepackage{longtable}
\usepackage{array}

\begin{document}

\title[A Review of Deep Transfer Learning and Recent Advancements]{A Review of Deep Transfer Learning and Recent Advancements}


\author*[1]{\fnm{Mohammadreza} \sur{Iman}}\email{m.iman@uga.edu}

\author[2]{\fnm{Khaled} \sur{Rasheed}}\email{khaled@uga.edu}

\author[1]{\fnm{Hamid Reza} \sur{Arabnia}}\email{hra@uga.edu}

\affil*[1]{\orgdiv{Department of Computer Science, Franklin College of Arts and Sciences}, \orgname{University of Georgia}, \orgaddress{\city{Athens}, \state{GA}, \country{USA}}}

\affil[2]{\orgdiv{Institute for Artificial Intelligence, Franklin College of Arts and Sciences}, \orgname{University of Georgia}, \orgaddress{\city{Athens}, \state{GA}, \country{USA}}}


\abstract{Deep learning has been the answer to many machine learning problems during the past two decades. However, it comes with two major constraints: dependency on extensive labeled data and training costs. Transfer learning in deep learning, known as Deep Transfer Learning (DTL), attempts to reduce such dependency and costs by reusing an obtained knowledge from a source data/task in training on a target data/task. Most applied DTL techniques are network/model-based approaches. These methods reduce the dependency of deep learning models on extensive training data and drastically decrease training costs. As a result, researchers detected Covid-19 infection on chest X-Rays with high accuracy at the beginning of the pandemic with minimal data using DTL techniques. Also, the training cost reduction makes DTL viable on edge devices with limited resources. Like any new advancement, DTL methods have their own limitations, and a successful transfer depends on some adjustments for different scenarios. In this paper, we review the definition and taxonomy of deep transfer learning and well-known methods. Then we investigate the DTL approaches by reviewing recent applied DTL techniques in the past five years. Further, we review some experimental analyses of DTLs to learn the best practice for applying DTL in different scenarios. Moreover, the limitations of DTLs (catastrophic forgetting dilemma and overly biased pre-trained models) are discussed, along with possible solutions and research trends.}




\maketitle

\section{Introduction}\label{sec1}

In recent years, Deep Learning (DL) has successfully addressed a number of challenging and interesting applications; in particular, problems that involved non-linearity of datasets. Recent advancements in deep learning methods deliver various usages and applications in extremely different areas such as image processing, natural language processing (NLP), numerical data analysis and predictions, and voice recognition. However, deep learning comes with restrictions, such as expensive training processes (time and processing) and the requirement of extensive training data (labeled data) \cite{1}. 

Since the start of the Machine Learning (ML) era, transfer learning has been a neat exploration for scientists. Before the rise of deep learning models, transfer learning was known as domain adaptation and focused on homogeneous data sets and how to relate such sets to each other because of the nature of ML algorithms \cite{2,3}. Traditional ML models have less dependency on dataset size, and usually, their training is less costly than deep learning models since they have been mostly designed for linear problems. Therefore, the motivation for using transfer learning in deep learning is higher than ever in the AI (Artificial Intelligence) and ML fields since it can address the two restraints of extensive training data and training costs.

Recent transfer learning methods on deep learning aim to reduce training process time and cost, and the necessity of extensive training datasets which can be hard to harvest in some areas such as medical images. Moreover, a pre-trained model for a specific job can be run on a simple edge device like a cellphone with limited processing capacity and limited training time \cite{4}. Also, developments in DTL are opening the door to more intuitive and sophisticated AI systems since it considers learning a continuous task. A great example of this idea is Google’s deep mind project and advancements such as progressive learning \cite{5}. All this is bringing DTL to the forefront of research in artificial intelligence and machine learning.

In this paper, first, the definition of DTL is reviewed, followed by the taxonomy of DTL. Then, selected recent practical studies of DTL are listed, categorized, and summarized. Moreover, two experimental evaluations of DTL and their conclusions are reviewed. Last but not least, we discuss the limitations of today's DTL techniques and possible ways to tackle them.

\section{Deep Learning}\label{sec2}

Deep learning (DL) or deep neural network (DNN) is a machine learning subcategory, which can deal with nonlinear datasets. DNNs consist of layers of stacked nodes, with activation function and associated weights, (fully/partially) connected and usually trained (weight adjustments) by back-propagation and optimization algorithms. During the past two decades, DNNs were developed rapidly and are used in many aspects of our daily lives today. For instance, Convolutional Neural Network (CNN) layers have improved deep learning models for visual-related tasks since 2011, and as of today, most DLs use CNN layers \cite{1}. For more details about machine learning and deep learning, please refer to \cite{1} since this paper is focused on deep transfer learning, and we assume that the reader should have a thorough understanding of machine learning and deep learning. 

\section{Deep Transfer Learning (DTL)}\label{sec3}

Deep transfer learning is about using the obtained knowledge from another task and dataset (even one not strongly related to the source task or dataset) to reduce learning costs. In many ML problems arranging a large amount of labeled data is impossible, which is mandatory for most DL models. For instance, at the beginning of the Covid-19 pandemic or even a year into it, providing enough chest X-Ray labeled data for training a deep learning model was still challenging, while using deep transfer learning, the AI achieved detecting the disease with very high accuracy with a limited training set \cite{13.1,13.2}. Another application is applying machine learning on edge devices such as phones for variant tasks by taking advantage of deep transfer learning to reduce the need for processing power.

An untrained DL uses a random initializing weight for nodes, and during the expensive training process, those weights adjust to the most optimized values by applying an optimization algorithm for a specific task (dataset). Remarkably, \cite{8} proved that initializing those weights based on a trained network with even a very distant dataset improves training performance compared to the random initialization. 

Deep transfer learning differs from semi-supervised learning since, in DTL, the source and target datasets can have a different distribution and just be related to each other, while in semi-supervised learning, the source and target data are from the same dataset, only the target set does not have the labels \cite{2}. DTL is also not the same as Multiview learning since Multiview learning uses two or more distinct datasets to improve the quality of one task, e.g., video datasets can be separated into image and audio datasets \cite{2}. Last but not least, DTL differs from Multitask learning despite many shared similarities. The most fundamental difference is that in Multitask learning, the tasks use interconnections to boost each other, and knowledge transfer happens concurrently between related tasks. In contrast in DTL, the target domain is the focus, and the knowledge has already been obtained for target data from source data, and they do not need to be related or function simultaneously \cite{2}.

\section{From Transfer Learning to Deep Transfer Learning, Taxonomy}\label{sec4}

It is possible to categorize Deep Transfer Learnings (DTLs) in different ways by various criteria, similar to Transfer Learnings. DTLs can be divided into two categories of homogeneous and heterogenous based on the homogeneity of source and target data \cite{2}. However, this categorization can be done differently because it is subjective and relative. For example, a dataset of X-Ray photos can be considered heterogeneous to a dataset of tree species photos when the comparison domain is limited to only image data. In contrast, it can be considered homogeneous to the same tree species photo dataset when the domain consists of audio and text datasets. 

Also, DTLs can be categorized into three groups based on label-setting aspects: (i) transductive, (ii) inductive, and (iii) unsupervised \cite{2}. Briefly, transductive is when only the source data is labeled; if both source and target data are labeled it is inductive; if none of the data are labeled it is unsupervised deep transfer learning \cite{2}.  

\cite{2} and \cite{9} mention and define another categorization of DTLs through the aspect of applied approaches. They similarly categorized DTLs into four groups of: (i) instance-based, (ii) feature-based / mapping-based, (iii) parameter-based / network-based, and (iv) relational-based / adversarial-based approaches. Instance-based transfer learning approaches are based on using selected parts of instances (or all) in source data and applying different weighting strategies to be used with target data. Feature-based approaches map instances (or some features) from both source and target data into more homogeneous data. Further, the \cite{2} survey divides the feature-based category into asymmetric and symmetric feature-based transfer learning subcategories. “Asymmetric approaches transform the source features to match the target ones. In contrast, symmetric approaches attempt to find a common latent feature space and then transform both the source and the target features into a new feature representation.” \cite{2} The network-based (parameter-based) methods are about using the obtained knowledge in the model (network) with different combinations of pre-trained layers: freezing some and/or finetuning some and/or adding some fresh layers. Relational/adversarial-based approaches focus on extracting transferable features from both source and target data either using the logical relationship or rules learned in the source domain or by applying methods inspired by generative adversarial networks (GAN) \cite{2,9}. Figure 1 shows the taxonomy of the above-mentioned categories \cite{2}.

\begin{figure}[h]%
\centering
\includegraphics[width=0.95\textwidth]{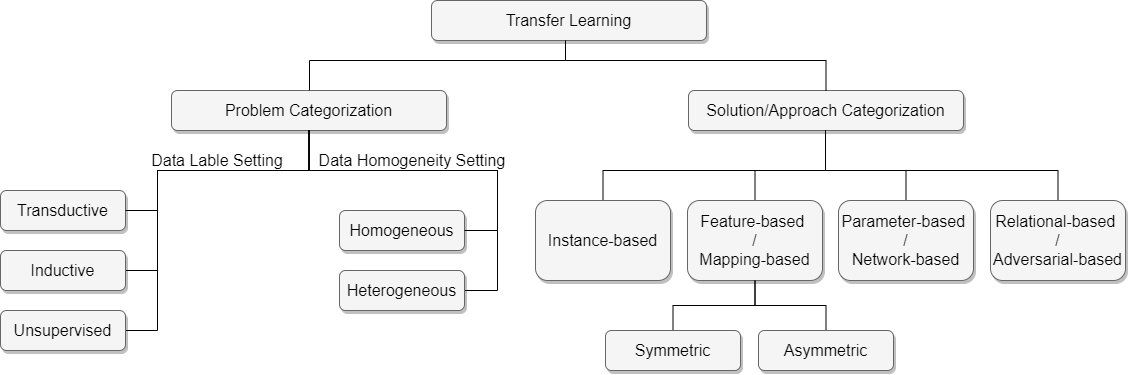}
\caption{Taxonomy of Transfer Learning which is extendable to Deep Transfer Learning as well.}\label{fig1}
\end{figure}

Other than the network-based and adversarial-based approaches, all other categories have been explored deeply during the last couple of decades for different ML techniques known as domain adaptation or transfer learning \cite{2,3}. However, most of those techniques are still applicable to deep transfer learning (DTL) as well. Network-based (parameter-based) approaches are the most applied techniques in DTL since they can tackle the domain adaptation between source and target data by adjusting the network (model). In other words, deep transfer learning is mainly focused on network-based approaches. Remarkably, network-based approaches in deep learning models can even tackle the adaptation of a very distant source and target data \cite{2,9}.

In deep transfer learning (DTL), different techniques are applied for network-based approaches, although generally, they are combinations of pre-training, freezing, finetuning, and/or adding a fresh layer(s). A deep learning network (DL model) trained on source data is called a pre-trained model consisting of pre-trained layers. Freezing and finetuning are techniques using some or all layers of pre-trained models to train the model on target data. Freezing some layers means the parameters/weights will not change and are constant values for frozen layers from a pre-trained model. finetune means the parameters/weights are initialized with the pre-trained values instead of random initialization for the whole network or some selected layers. Another recent DTL technique is based on freezing a pre-trained model and adding new layers to that model for training on target data; Google's deep mind project introduces this technique in 2016 as Progressive Learning / progressive neural networks (PNNs) \cite{5,10}. 

The concept of progressive learning mimics human skill learning, which is adding a new skill on top of previously learned skills as a foundation to learn a new one. E.g., a child learns how to run after learning to crawl and walk and using all the skills obtained in the process. Similarly, PNNs prevent catastrophic forgetting in DTL versus finetuning techniques by freezing the whole pre-trained model and learning (adjusting to) the new task by training the newly added layers on top of previously trained layers \cite{5,10}.

In deep learning models, usually, the earlier layers do the feature extraction at a high level of detail, further layers towards the end extract the information and conceptualize the given data, and lateral layers do the classifications or predictions. For instance, in the image-related model, the earlier layers of CNN extract the edges, corners, and tiny patches of a given image. Further layers put those details together to detect objects or faces, and the lateral layers, usually fully connected layers, do the classification \cite{11}. Given this process, the most effective and efficient approach for DTL, to our knowledge, is to freeze the earlier and middle layers from a related pre-trained model and finetune the lateral layers for the new task/dataset \cite{12}. Similarly, the new layers are added to the last part of a pre-trained model in progressive learning. 

Nonetheless, some other research in this area use combinational and sophisticated methods to tackle transfer learning in deep learning like ensembled networks, weighting strategies, etc. \cite{2}. However, to our knowledge, the search for recent advancements in DTL for practical tasks ends up with methods based on mostly the network-based and limited number of adversarial-based approaches.

\section{Review of Recent Advancements in DTL}\label{sec5}

We limited our selection to the last five years of published studies on deep transfer learning for various tasks and data types. Table 1 shows the list of selected works from hundreds of reviewed literature sorted by their DTL approaches. We used the systematic literature review (SLR) technique \cite{13} for the process of finding and selecting these thirty-eight publications. The inclusion criteria that we used for our selection process are as follows: a) published in the past five years, b) reproducible (detailed implementation and models), c) applied to practical ML problems, and d) generalizable. We found that all reviewed studies mostly fall into three categories of network-based approaches and some into the adversarial-based approach, which are explained in the previous section. We name these approaches as (i) Finetuning: finetuning a pre-trained model on target data; (ii) Freezing CNN layers: the earlier CNN layers are frozen, and only the lateral fully connected layers are finetuned; (iii) Progressive learning: some or all layers of a pre-trained model are selected and used frozen, and some fresh layers will be added to the model to be trained on target data; and (iv) Adversarial-based: extracting transferable features from both source and target data using adversarial or relational methods, Figure 2.

\begin{figure}[h]%
\centering
\includegraphics[width=0.8\textwidth]{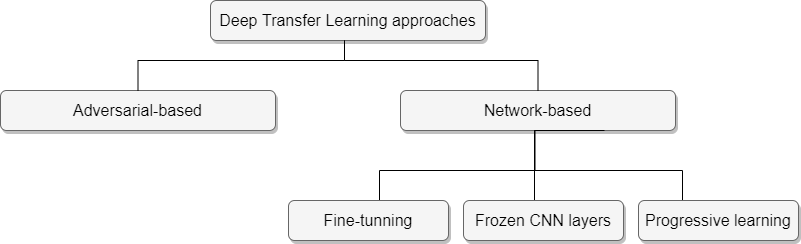}
\caption{Most common Deep Transfer Learning approaches.}\label{fig2}
\end{figure}

The most common DTL method is using a trained model on a highly related dataset to target data and finetune it on target data (finetuning). The simplicity of applying this technique makes it the most popular DTL method in our selection; 21 of 38 selected works have used this method. This method can improve training on target data in various ways, such as reducing training costs and tackling the need for an extensive target dataset. However, it is still prone to catastrophic forgetting. Needless to say, it is a very effective DTL method for many tasks and datasets in various fields such as medical, mechanics, art, physics, security, etc. Also, it has been applied for both image datasets and tabular (numerical) datasets as listed in Table 1.

The second popular approach in DTL is freezing CNN layers in a pre-trained model and finetune only lateral fully connected layers (Freezing CNN layers). CNN layers extract features from the given dataset, and the fully connected layers are responsible for classification, which in this method will be finetuned to the new task for target data.

\setlength{\tabcolsep}{1pt}
\tiny
\begin{longtable}{| m{0.035\linewidth} | m{0.045\linewidth}  | m{0.3\linewidth}  | m{0.08\linewidth}  | m{0.06\linewidth}  | m{0.11\linewidth}  | m{0.05\linewidth}  | m{0.12\linewidth}  | m{0.13\linewidth}  |}
\caption{List of selected recent deep transfer learning (DTL) publications.\label{long}}\\

\hline
\multicolumn{9}{| c |}{Begin of Table}\\
\hline
\textbf{Ref.} & \textbf{Year} & \textbf{Title} & \textbf{Data Type} & \textbf{Time Series} & \textbf{Approach} & \textbf{CNN} & \textbf{Known Models Used} & \textbf{Dataset Field} \\
\hline
\endfirsthead

\hline
\multicolumn{9}{|c|}{Continuation of Table \ref{long}}\\
\hline
\textbf{Ref.} & \textbf{Year} & \textbf{Title} & \textbf{Data Type} & \textbf{Time Series} & \textbf{Approach} & \textbf{CNN} & \textbf{Known Models Used} & \textbf{Dataset Field} \\
\hline
\endhead

\hline
\endfoot

\hline
\multicolumn{9}{| c |}{End of Table}\\
\hline\hline
\endlastfoot

\cite{11.01}&2022&UAV swarm based radar signal sorting via multi-source data fusion: A deep transfer learning framework&Image&no&Finetuning&yes&Yolo, Faster-RCNN, Cascade-RCNN&Radar image  \\
\hline
\cite{11.1}&2022&Classification of analyzable metaphase images using transfer learning and fine tuning&Image&no&Finetuning&yes&VGG16, Inception V3&Medical image  \\
\hline
\cite{11.2}&2021&Multiclassification of Endoscopic Colonoscopy Images Based on Deep Transfer Learning&Image&no&Finetuning&yes&AlexNet, VGG, Res-Net&Medical Image  \\
\hline
\cite{11.3}&2021&MCFT-CNN: Malware classification with fine-tune convolution neural networks using traditional and transfer learning in Internet of Things&Image&no&Finetuning&yes&Res-Net50&Malware classification  \\
\hline
\cite{11.4}&2021&Facial Emotion Recognition Using Transfer Learning in the Deep CNN&Image&no&Finetuning&yes&VGGs, Res-Nets, Inception-v3, DenseNet-161&Facial emotion recognition (FER)  \\
\hline
\cite{13.1}&2020&Automated Deep Transfer Learning-Based Approach for Detection of COVID-19 Infection in Chest X-rays&Image&no&Finetuning&yes&Inception-Xception&Medical image  \\
\hline
\cite{13.2}&2020&Classification of the COVID-19 infected patients using DenseNet201 based deep transfer learning&Image&no&Finetuning&yes&ImageNet, Dense-Net&Medical image  \\
\hline
\cite{14}&2019&	Enhancing materials property prediction by leveraging computational and experimental data using deep transfer learning&	Tabular/ bigdata&	yes&	Finetuning&	no&	none&	Quantum mechanics \\
\hline
\cite{15}&
2019&	Application of deep transfer learning for automated brain abnormality classification using MR images&	Image&	no&	Finetuning&	yes& Res-Net&	Medical image \\
\hline
\cite{16}&
2019&	An adaptive deep transfer learning method for bearing fault diagnosis&	Tabular/ bigdata&	yes&	Finetuning&	no&	LSTM RNN&	Mechanic \\
\hline
\cite{17}&
2019&	Online detection for bearing incipient fault based on deep transfer learning&	Image&	yes&	Finetuning&	yes&	VGG-16&	Mechanic \\
\hline
\cite{18}&
2019&	Towards More Accurate Automatic Sleep Staging via Deep Transfer Learning&	Tabular/ bigdata&	yes&	Finetuning&	yes&	none&	Medical data \\
\hline
\cite{19}&
2019&	Deep Transfer Learning for Multiple Class Novelty Detection&	Image&	no&	Finetuning&	yes&	Alex-Net, VGG-Net&	Vision \\
\hline
\cite{20}&
2019&	A Digital-Twin-Assisted Fault Diagnosis Using Deep Transfer Learning&	Tabular/ bigdata&	no&	Finetuning&	no&	none&	Mechanic \\
\hline
\cite{21}&
2019&	Learning to Discover Novel Visual Categories via Deep Transfer Clustering&	Image&	no&	Finetuning&	yes&	none&	Vision \\
\hline
\cite{22}&
2018&	Deep Transfer Learning for Person Re-identification&	Image&	no&	Finetuning&	yes&	none&	Identification / security \\
\hline
\cite{23}&
2018&	Deep Transfer Learning for Art Classification Problems&	Image&	no&	Finetuning&	yes&	none&	Art \\

\hline
\cite{24}&
2018&	Classification and unsupervised clustering of LIGO data with Deep Transfer Learning&	Image&	no&	Finetuning&	yes&	none&	Physics / Astrophysics \\
\hline
\cite{25}&
2018&	Empirical Study and Improvement on Deep Transfer Learning for Human Activity Recognition&	Tabular/ bigdata&	yes&	Finetuning&	yes&	none&	Human Activity Recognition \\
\hline
\cite{26}&
2018&	Automatic ICD-9 coding via deep transfer learning&	Tabular/ bigdata&	no&	Finetuning&	yes&	none&	Medical \\
\hline
\cite{27}&
2017&	Video-based emotion recognition in the wild using deep transfer learning and score fusion&	video (audio \& visual)&	yes&	Finetuning&	yes&	VGG-Face&Human science / psychology \\
\hline
\cite{27.1}&
2022&	Deep transfer learning-based visual classification of pressure injuries stages&	Image&	no&	Freezing CNN layers&	yes&	Dense-Net 121, Inception V3, MobilNet V2, Res-Nets, VGG16&	Medical image \\
\hline
\cite{27.2}&
2021&	Deep Transfer Learning for WiFi Localization&	Tabular/ bigdata&	no&	Freezing CNN layers&	yes&	none&	WiFi Localization \\
\hline
\cite{28}&
2020&	Automated invasive ductal carcinoma detection based using deep transfer learning with whole-slide images&	Image&	no&	Freezing CNN layers&	yes&	Res-Net, Dense-Net&	Medical image \\
\hline
\cite{29}&
2019&	Deep Transfer Learning for Signal Detection in Ambient Backscatter Communications&	Tabular/ bigdata&	no&	Freezing CNN layers&	yes&	none&	Tele-communication \\
\hline
\cite{30}&
2019&	Brain tumor classification using deep CNN features via transfer learning&	Image&	no&	Freezing CNN layers&	yes&	Google-Net&	Medical image \\

\hline
\cite{31}&
2018&	Comparison of Deep Transfer Learning Strategies for Digital Pathology&	Image&	no&	Freezing CNN layers&	yes&	none&	Medical image \\
\hline
\cite{32}&
2018&	Deep transfer learning for military object recognition under small training set condition&	Image&	no&	Freezing CNN layers&	yes&	none&	Military \\
\hline
\cite{33}&
2018&	Deep Transfer Learning for Image‐Based Structural Damage Recognition&	Image&	no&	Freezing CNN layers&	yes&	VGG-Net&	Civil engineering \\
\hline
\cite{34}&
2017&	Deep Transfer Learning for Modality Classification of Medical Images&	Image&	no&	Freezing CNN layers&	yes&	VGG-Net, Res-Net&	Medical image \\
\hline
\cite{35}&
2017&	Folding Membrane Proteins by Deep Transfer Learning&	Tabular/ bigdata&	no&	Freezing CNN layers&	yes&	Res-Net&	Chemistry \\
\hline
\cite{35.2}&
2021&	Progressive Transfer Learning Approach for Identifying the Leaf Type by Optimizing Network Parameters&	Image&	no&	Progressive learning&	yes&	Res-Net50&	Plant science \\
\hline
\cite{36}&
2020&	An Evaluation of Progressive Neural Networks for Transfer Learning in Natural Language Processing&	NLP / text&	no&	Progressive learning&	no&	none&	NLP \\
\hline
\cite{37}&
2020&	Progressive Transfer Learning and Adversarial Domain Adaptation for Cross-Domain Skin Disease Classification&	Image&	no&	Progressive learning&	yes&	none&	Medical image \\
\hline
\cite{38}&
2017&	Progressive Neural Networks for Transfer Learning in Emotion Recognition&	Image \& audio&	yes&	Progressive learning&	no&	none&	Para-linguistic \\
\hline
\cite{39}&
2020&	A deep transfer learning model with classical data augmentation and CGAN to detect COVID-19 from chest CT radiography digital images&	Image&	no&	Adversarial-based&	yes&	Alex-Net, VGG-Net16, VGG-Net19, Google-Net, Res-Net50&	Medical image \\
\hline
\cite{40}&
2019&	Diagnosing Rotating Machines with Weakly Supervised Data Using Deep Transfer Learning&	Tabular/ bigdata&	yes&	Adversarial-based&	yes&	none&	Mechanic \\
\hline
\cite{41}&
2017&	A New Deep Transfer Learning Based on Sparse Auto-Encoder for Fault Diagnosis&	Tabular/ bigdata&	yes&	Sparse Auto-Encoder&	no&	none&	Mechanic \\

\end{longtable}
\normalsize

\cite{27.1,27.2,28,29,30,31,32,33,34,35} are the sample research publications, which have used this method for different data types such as image and tabular data as listed in Table 1. This technique is specific to the models consisting of CNN layers; however, it can be extended to other deep learning models by assuming the earlier and middle layers are acting similar to CNN layers for feature extraction.

Using well-known models such as VGG-Net, Alex-Net, and Res-Net, which has already been trained on ImageNet datasets \cite{42}, is a general approach for both of the techniques mentioned above since they are easily accessible, and they are pre-trained to the highest possible accuracy. It is worth mentioning that such training can take days of processing time even with clusters of GPUs/TPUs and the mentioned methods are skipping the pre-training step by simply downloading a publicly available pre-trained model.

\cite{35.2,36,37,38} are based on the progressive learning method, also known as progressive neural networks (PNNs), described earlier. \cite{36} evaluates progressive learning effectiveness for common natural language processing (NLP) tasks: sequence labeling and text classification. Through evaluation and comparison of applying PNNs to various models, datasets, and tasks, they show how PNNs improve DL models' accuracy by avoiding catastrophic forgetting in finetuning techniques. \cite{35.2,37,38} use PNNs for image and audio datasets and similarly finds tangible improvements in comparison to other DTL techniques.

\cite{39} and \cite{40} are examples of adversarial-based approaches that we found in the literature. In \cite{39}, they used conditional generative adversarial networks (CGAN) to expand limited target data of chest X-Ray images for detecting Covid-19 DTL model. \cite{40} applies domain adversarial training to obtain the shared features between multiple source datasets.

Moreover, we found some tailored DTL methods for specific tasks and datasets like \cite{41}. The proposed method in \cite{41} as they describe is based on “three-layer sparse auto-encoder to extract the features of raw data, and applies the maximum mean discrepancy term to minimizing the discrepancy penalty between the features from training data and target data.” They tailor that method for smart industry fault diagnosis problems and achieve 99.82\% accuracy which is better than other approaches like deep belief network, sparse filter, deep learning, and support vector machine. Such tailored DTL approaches are not usually easy to generalize for different tasks or datasets. Nonetheless, they can open the door to interesting and new techniques in deep transfer learning’s future.

\section{Experimental Analyzations of Deep Transfer Learning}\label{sec6}

In this section we review two remarkable experimental evaluations of DTL techniques. The tests’ setup, analysis, and conclusions are noteworthy for applying DTL techniques in different scenarios.

“What is being transferred in transfer learning?” \cite{43} is a recent experimental study which uses a series of tests on visual domain and deep learning models and tries to investigate what makes a successful transfer and which part of the network is responsible for that. To do so, they analyze networks in four different cases: (i) pre-trained network, (ii) random initialized network, (iii) finetuned network on target domain after pretraining on source domain, (iv) trained network from random initialization \cite{43}. Moreover, to characterize the role of feature reuse, they use a source (pre-train) domain containing natural images (IMAGENET), and a few target (downstream) domains with decreasing visual similarities from natural images: DOMAINNET real, DOMAINNET clipart, CHEXPERT (medical chest X-Rays) and DOMAINNET quickdraw \cite{43}.

The study shows that feature reuse plays a key role in deep transfer learning as a pre-trained model on IMAGENET shows the largest performance improvement on real domain, which shares similar visual features (natural images) with IMAGENET in comparison to randomly initialized models. Also, they run a series of experiments by shuffling the image blocks (different block sizes). These experiments prove that feature reuse plays a very important role in transfer learning, particularly when the target domain shares visual features with the source domain. However, they realize that feature reuse is not the only reason for deep transfer learning success since even for distant targets such as CHEXPERT and quickdraw, they still observe performance boosts from deep transfer learning. Additionally, in all cases pre-trained models converge way faster than random initialized models. \cite{43}

Further, they manually analyze common and uncommon mistakes in the training of randomly initialized versus pre-trained models. They observe that data samples marked incorrect in the pre-trained model and correct in the randomly initialized model are mostly ambiguous samples. On the other hand, the majority of the samples that a pre-trained model marked correct and a randomly initialized model marked incorrect are straightforward samples. This means that a pre-trained model has a stronger prior, and it is harder to adapt to the target domain. Moreover, using centered kernel alignment to measure feature similarities, they conclude that the initialization point drastically impacts feature similarity, and two networks with high accuracy can have a different feature space. Also, they discover similar results for distance in parameter space, which two random-initialized models are farther from each other compared to two pre-trained models. \cite{43}

In regard to performance barriers and basins in the loss landscape, they have concluded that the network stays in the same basin of solution when finetuning a pre-trained network. They reach to this conclusion by training pre-trained models from two random runs as well as training random initialized models twice and comparing. Even when training a random initialized model two times with the same random values the models end up in different basins. \cite{43}

Module criticality is an interesting analysis of deep learning models. Usually, in a deep CNN model each layer of CNN considers a module, while in some models a component of network can be considered as a module. To measure criticality of a module, it is possible to take a trained model and re-initialize each module at once and compare the amount of model accuracy drop. Adopting this technique, the authors of \cite{43} discovered: (i) fully connected layers (near to model output) become critical for P-T model, and (ii) module criticality increases moving from the input side of model towards output, which is consistent with the concept of earlier layers (near input) extracting more general features while lateral layers have features that are more specialized for the target domain.

\cite{44} is another experimental analysis of transfer learning in visual tasks with the title of “Factors of Influence for Transfer Learning across Diverse Appearance Domains and Task Types”. Three factors of influence are investigated in this study: (i) image domain, the difference in image domain between source and target tasks, (ii) task type, the difference in task type, and (iii) dataset size, the size of the source and target training sets. They perform over 1200 transfer learning experiments on 20 datasets spanning seven diverse image domains (consumer, driving, aerial, underwater, indoor, synthetic, closeups) and four task types (semantic segmentation, object detection, depth estimation, keypoint detection). \cite{44}

They use data normalization (e.g., Illumination normalization) and augmentation techniques to improve models’ accuracy. They adopt recent high-resolution backbone HRNetV2, which consists of 69M parameters. This backbone is easily adjustable for different datasets by simply replacing the head of the backbone. To make a fair comparison they pre-trained (to be used for transfer learning) their models from scratch and evaluated their performance using top-1 accuracy on the ILSVRC’12 validation set. \cite{44}

The transfer learning experiments are mainly divided into two settings of (i) transfer learning with small target training set and (ii) with the full target set. The evaluation of transfer learning models is based on the gain obtained from finetuning from a specific source model compared to finetuning from ILSVRC’12 image classification with the main question of “are additional gains possible, by picking a good source?”. Furthermore, they added a series of experiments for multi-source training to investigate the impact of using multi-source training for a specific task. \cite{44}

Such an exhaustive experimental analysis resulted in following observations: (i) all experiments proved that transfer learning outperforms training from scratch (random initialization); (ii) for 85\% of target tasks there exists a source task which tops ILSVCR’12 pre-training; (iii) the most transfer gain happens when the source and target tasks are in the same image domain (within-domain), which is even more important than source size; (iv) positive transfer gain is possible when the source image domain includes the target domain; (v) although multisource models bring good transfer, they are outperformed by the largest within-domain source; (vi) “for 65\% of the targets within the same image domain as the source, cross-task-type transfer results in positive transfer gains”; (vii) as naturally expected, the larger datasets positively transfer towards the smaller datasets; (viii) transfer effects are stronger for a small target training set, which helps the process of choosing the transfer learning model by testing several models with a small section of target data. \cite{44}

\section{Discussion}\label{sec7}

The Deep Transfer Learning (DTL) research field is thriving because of the motivation to handle the limitations of Deep Learning (DL) models, which are the dependency on extensive labeled data and training costs. The main idea is to use obtained knowledge from source data in the training process on target data. Another possible impactful outcome of the DTL research line is to achieve continual learning, which brings Artificial General Intelligence \cite{1} a step closer to reality. Continual learning can be achieved simply through a chain of transfer learning processes while the end model is still valid on all previous training sources.

As we reviewed in previous sections, model-based approaches are the most commonly used approaches in DTL since deep learning models have the capacity to be adjusted to transfer knowledge. However, there are two main constraints in such approaches— catastrophic forgetting dilemma and an overly biased pre-trained model.

In the case of finetuning a pre-trained model, there is a high chance of drastic changes of weights through the whole model resulting in the catastrophic forgetting dilemma. Therefore, the obtained knowledge could be partially or even completely wiped out, resulting in unsuccessful training and no possibility of continual learning. This constraint limits the success of the finetuning approach to tightly related source and target data. Also, a very well-known technique to reduce the forgetting effect is to add a limited number of source samples to the target training data.

Freezing the pre-trained CNN layers technique tries to tackle the catastrophic forgetting by freezing the obtained knowledge on earlier layers and finetuning the fully-connected lateral layers to achieve transfer learning for target data. Given the fact that earlier layers in DL models extract detailed features and move towards the output, more abstract knowledge is extracted \cite{11}; freezing the earlier layers limits the ability of the model to learn any new features from target data, which is known as an overly biased pre-trained model. Having extensive source data or access to a pre-trained model on a large dataset is critical for a successful transfer using this technique. In this way, there is a high chance that the pre-trained model has already learned any possible detailed features, and simply by finetuning the lateral layers can perform on target data. However, even tackling the first obstacle, this solution is still imperiled by the catastrophic forgetting in lateral layers. This technique is still successful in the case of related source and target data and tasks despite the limitations mentioned above.

Progressive learning tries to find a middle ground between catastrophic forgetting and a biased model by adding a new layer(s) to the end of a frozen pre-trained model. This technique is successful in the case of task transfer for related source and target data. It can not deal with distant source and target data since the earlier layers are frozen and cannot learn new features; however, the new lateral layer helps the model adjust to a new task.

A possible solution to address both catastrophic forgetting and an overly biased pre-trained model in DTL is to increase the learning capacity of a pre-trained model by vertically expanding it. In another research paper we propose expanding the model vertically in training on target data, adding new nodes on frozen pre-trained layers throughout the model instead of adding a new layer(s) to the end of the model \cite{45}. The vertical expansion increases the model learning capacity while keeping the previously obtained knowledge intact. Therefore, not only do we achieve successful transfer learning, our final model is still valid on source data opening the door to deep continual learning. \cite{45}

\section{Conclusion}\label{sec8}

This paper reviews the taxonomy of deep transfer learning (DTL) and the definitions of different approaches. Also, we review, list, categorize and analyze over thirty recent applied DTL research studies. Then, we investigate the methodology and limitations of the three most common model-based deep transfer learning methods: (i) Finetuning, (ii) Freezing CNN Layers, and (iii) Progressive Learning. These techniques have proven their ability and effectiveness for various machine learning problems. The simplicity of finetuning publicly available pre-trained models on extensive datasets is the reason for it being the most common transfer learning technique. Moreover, two thorough experimental studies in DTL are summarized; their discoveries clarify the details of a successful deep transfer learning approach for different scenarios. Last but not least, the limitations of current DTLs, catastrophic forgetting dilemma, and overly biased pre-trained models are discussed, along with possible solutions.


\begin{thebibliography}{44} 
\addtolength{\leftmargin}{0.1in}
\bibitem{1} M. Iman, H. R. Arabnia, and R. M. Branchinst, “Pathways to Artificial General Intelligence: A Brief Overview of Developments and Ethical Issues via Artificial Intelligence, Machine Learning, Deep Learning, and Data Science,” Springer, Cham, pp. 73–87, 2021.
\bibitem{2}	Zhuang, Fuzhen, Zhiyuan Qi, Keyu Duan, Dongbo Xi, Yongchun Zhu, Hengshu Zhu, Hui Xiong, and Qing He. "A comprehensive survey on transfer learning." Proceedings of the IEEE 109, no. 1, 43-76, 2020.
\bibitem{3}	Farahani, Abolfazl, Sahar Voghoei, Khaled Rasheed, and Hamid R. Arabnia. "A brief review of domain adaptation." Advances in Data Science and Information Engineering, 877-894, 2021.
\bibitem{4}	S. Voghoei, N. Hashemi Tonekaboni, J. G. Wallace, and H. R. Arabnia, “Deep learning at the edge,” Proc. - 2018 Int. Conf. Comput. Sci. Comput. Intell. CSCI 2018, pp. 895–901, Dec. 2018.
\bibitem{5}	H. S. Chang, M. C. Fu, J. Hu, and S. I. Marcus, “Google Deep Mind’s AlphaGo,” OR/MS Today, vol. 43, no. 5, pp. 24–29, 2016.
\bibitem{8}	J. Yosinski, J. Clune, Y. Bengio, and H. Lipson, “How transferable are features in deep neural networks?,” Adv. Neural Inf. Process. Syst., vol. 4, no. January, pp. 3320–3328, Nov. 2014.
\bibitem{9}	C. Tan, F. Sun, T. Kong, W. Zhang, C. Yang, and C. Liu, “A survey on deep transfer learning,” Lect. Notes Comput. Sci. (including Subser. Lect. Notes Artif. Intell. Lect. Notes Bioinformatics), vol. 11141 LNCS, pp. 270–279, 2018.
\bibitem{10}	Rusu, Andrei A., Neil C. Rabinowitz, Guillaume Desjardins, Hubert Soyer, James Kirkpatrick, Koray Kavukcuoglu, Razvan Pascanu, and Raia Hadsell. "Progressive neural networks." arXiv preprint arXiv:1606.04671 (2016).
\bibitem{11}	J. Yosinski, J. Clune, A. Nguyen, T. Fuchs, and H. Lipson, “Understanding neural networks through deep visualization,” arXiv Prepr. arXiv1506.06579, 2015.
\bibitem{12}	Ravishankar, Hariharan, Prasad Sudhakar, Rahul Venkataramani, Sheshadri Thiruvenkadam, Pavan Annangi, Narayanan Babu, and Vivek Vaidya. "Understanding the mechanisms of deep transfer learning for medical images." In Deep learning and data labeling for medical applications, pp. 188-196. Springer, Cham, 2016.
\bibitem{13}	B. Kitchenham, O. Pearlbrereton, D. Budgen, M. Turner, J. Bailey, and S. Linkman, “Systematic literature reviews in software engineering – A systematic literature review,” Inf. Softw. Technol., vol. 51, no. 1, pp. 7–15, Jan. 2009.
\bibitem{11.01} Wan, L., Liu, R., Sun, L., Nie, H. and Wang, X., “UAV swarm based radar signal sorting via multi-source data fusion: A deep transfer learning framework.”, Information Fusion, 78, pp.90-101, 2022.
\bibitem{11.1} Albayrak, A., “Classification of analyzable metaphase images using transfer learning and fine tuning.” Med Biol Eng Comput 60, 239–248, 2022. https://doi.org/10.1007/s11517-021-02474-z
\bibitem{11.2} Kumar, S., “MCFT-CNN: Malware classification with fine-tune convolution neural networks using traditional and transfer learning in internet of things. Future Generation Computer Systems”, 125, pp.334-351, 2021.
\bibitem{11.3} Wang, Y., Feng, Z., Song, L., Liu, X. and Liu, S.,  “Multiclassification of endoscopic colonoscopy images based on deep transfer learning.”, Computational and Mathematical Methods in Medicine, 2021.
\bibitem{11.4} Akhand, M.A.H., Roy, S., Siddique, N., Kamal, M.A.S. and Shimamura, T., “Facial Emotion Recognition Using Transfer Learning in the Deep CNN.”,  Electronics, 10(9), p.1036, 2021.
\bibitem{13.1}	N. Narayan Das, N. Kumar, M. Kaur, V. Kumar, and D. Singh, “Automated Deep Transfer Learning-Based Approach for Detection of COVID-19 Infection in Chest X-rays,” Irbm, vol. 1, pp. 1–6, 2020.
\bibitem{13.2}	A. Jaiswal, N. Gianchandani, D. Singh, V. Kumar, and M. Kaur, “Classification of the COVID-19 infected patients using DenseNet201 based deep transfer learning,” J. Biomol. Struct. Dyn., pp. 1–8, Jul. 2020.
\bibitem{14}	Jha, Dipendra, Kamal Choudhary, Francesca Tavazza, Wei-keng Liao, Alok Choudhary, Carelyn Campbell, and Ankit Agrawal. "Enhancing materials property prediction by leveraging computational and experimental data using deep transfer learning." Nature communications 10, no. 1 (2019): 1-12.
\bibitem{15}	M. Talo, U. B. Baloglu, Ö. Yıldırım, and U. Rajendra Acharya, “Application of deep transfer learning for automated brain abnormality classification using MR images,” Cogn. Syst. Res., vol. 54, pp. 176–188, May 2019.
\bibitem{16}	Z. Wu, H. Jiang, K. Zhao, and X. Li, “An adaptive deep transfer learning method for bearing fault diagnosis,” Measurement, vol. 151, p. 107227, Feb. 2020.
\bibitem{17}	W. Mao, L. Ding, S. Tian, and X. Liang, “Online detection for bearing incipient fault based on deep transfer learning,” Meas. J. Int. Meas. Confed., vol. 152, p. 107278, Feb. 2020.
\bibitem{18}	Phan, Huy, Oliver Y. Chén, Philipp Koch, Zongqing Lu, Ian McLoughlin, Alfred Mertins, and Maarten De Vos. "Towards more accurate automatic sleep staging via deep transfer learning." IEEE Transactions on Biomedical Engineering 68, no. 6 (2020): 1787-1798.
\bibitem{19}	P. Perera and V. M. Patel, “Deep Transfer Learning for Multiple Class Novelty Detection,” 2019.
\bibitem{20}	Y. Xu, Y. Sun, X. Liu, and Y. Zheng, “A Digital-Twin-Assisted Fault Diagnosis Using Deep Transfer Learning,” IEEE Access, vol. 7, pp. 19990–19999, 2019.
\bibitem{21}	K. Han, A. Vedaldi, and A. Zisserman, “Learning to Discover Novel Visual Categories via Deep Transfer Clustering,” 2019.
\bibitem{22}	M. Geng, Y. Wang, T. Xiang, and Y. Tian, “Deep Transfer Learning for Person Re-identification,” 2018 IEEE 4th Int. Conf. Multimed. Big Data, BigMM 2018, Nov. 2016.
\bibitem{23}	M. Sabatelli, M. Kestemont, W. Daelemans, and P. Geurts, “Deep Transfer Learning for Art Classification Problems,” 2018.
\bibitem{24}	D. George, H. Shen, and E. A. Huerta, “Deep transfer learning: A new deep learning glitch classification method for advanced ligo,” arXiv, vol. 97, no. 10. arXiv, p. 101501, 22-Jun-2017.
\bibitem{25}	Ding, Renjie, Xue Li, Lanshun Nie, Jiazhen Li, Xiandong Si, Dianhui Chu, Guozhong Liu, and Dechen Zhan. "Empirical study and improvement on deep transfer learning for human activity recognition." Sensors 19, no. 1 (2019): 57.
\bibitem{26}	M. Zeng, M. Li, Z. Fei, Y. Yu, Y. Pan, and J. Wang, “Automatic ICD-9 coding via deep transfer learning,” Neurocomputing, vol. 324, pp. 43–50, 2019.
\bibitem{27}	H. Kaya, F. Gürpınar, and A. A. Salah, “Video-based emotion recognition in the wild using deep transfer learning and score fusion,” Image Vis. Comput., vol. 65, pp. 66–75, Sep. 2017.
\bibitem{27.1}	Ay, B., Tasar, B., Utlu, Z., Ay, K. and Aydin, G., “Deep transfer learning-based visual classification of pressure injuries stages.”, Neural Computing and Applications, pp.1-12, 2022.
\bibitem{27.2}	Li, P., Cui, H., Khan, A., Raza, U., Piechocki, R., Doufexi, A. and Farnham, T., “May. Deep transfer learning for WiFi localization.”, In 2021 IEEE Radar Conference (RadarConf21) (pp. 1-5). IEEE., 2021.
\bibitem{28}	Y. Celik, M. Talo, O. Yildirim, M. Karabatak, and U. R. Acharya, “Automated invasive ductal carcinoma detection based using deep transfer learning with whole-slide images,” Pattern Recognit. Lett., vol. 133, pp. 232–239, May 2020.
\bibitem{29}	C. Liu, Z. Wei, D. W. K. Ng, J. Yuan, and Y. C. Liang, “Deep Transfer Learning for Signal Detection in Ambient Backscatter Communications,” IEEE Trans. Wirel. Commun., 2020.
\bibitem{30}	S. Deepak and P. M. Ameer, “Brain tumor classification using deep CNN features via transfer learning,” Comput. Biol. Med., vol. 111, no. June, p. 103345, 2019.
\bibitem{31}	R. Mormont, P. Geurts, and R. Marée, “Comparison of deep transfer learning strategies for digital pathology,” 2018.
\bibitem{32}	Yang, Zhi, Wei Yu, Pengwei Liang, Hanqi Guo, Likun Xia, Feng Zhang, Yong Ma, and Jiayi Ma. "Deep transfer learning for military object recognition under small training set condition." Neural Computing and Applications 31, no. 10 (2019): 6469-6478.
\bibitem{33}	Y. Gao and K. M. Mosalam, “Deep Transfer Learning for Image-Based Structural Damage Recognition,” Comput. Civ. Infrastruct. Eng., vol. 33, no. 9, pp. 748–768, Sep. 2018.
\bibitem{34}	Y. Yu, H. Lin, J. Meng, X. Wei, H. Guo, and Z. Zhao, “Deep Transfer Learning for Modality Classification of Medical Images,” Information, vol. 8, no. 3, p. 91, Jul. 2017.
\bibitem{35}	S. Wang, Z. Li, Y. Yu, and J. Xu, “Folding Membrane Proteins by Deep Transfer Learning,” Cell Syst., vol. 5, no. 3, pp. 202-211.e3, Sep. 2017.
\bibitem{35.2}	Joshi, D., Mishra, V., Srivastav, H. and Goel, D., “Progressive Transfer Learning Approach for Identifying the Leaf Type by Optimizing Network Parameters.”, Neural Processing Letters, 53(5), pp.3653-3676., 2021.
\bibitem{36}	Moeed, Abdul, Gerhard Hagerer, Sumit Dugar, Sarthak Gupta, Mainak Ghosh, Hannah Danner, Oliver Mitevski, Andreas Nawroth, and Georg Groh. "An evaluation of progressive neural networksfor transfer learning in natural language processing." In Proceedings of The 12th Language Resources and Evaluation Conference, pp. 1376-1381. 2020.
\bibitem{37}	Y. Gu, Z. Ge, C. P. Bonnington, and J. Zhou, “Progressive Transfer Learning and Adversarial Domain Adaptation for Cross-Domain Skin Disease Classification,” IEEE J. Biomed. Heal. Informatics, vol. 24, no. 5, pp. 1379–1393, May 2020.
\bibitem{38}	J. Gideon, S. Khorram, Z. Aldeneh, D. Dimitriadis, and E. M. Provost, “Progressive Neural Networks for Transfer Learning in Emotion Recognition,” Proc. Annu. Conf. Int. Speech Commun. Assoc. INTERSPEECH, vol. 2017-August, pp. 1098–1102, Jun. 2017.
\bibitem{39}	M. Loey, G. Manogaran, and N. E. M. Khalifa, “A deep transfer learning model with classical data augmentation and CGAN to detect COVID-19 from chest CT radiography digital images,” Neural Comput. Appl., pp. 1–13, Oct. 2020.
\bibitem{40}	X. Li, W. Zhang, Q. Ding, and X. Li, “Diagnosing Rotating Machines with Weakly Supervised Data Using Deep Transfer Learning,” IEEE Trans. Ind. Informatics, vol. 16, no. 3, pp. 1688–1697, Mar. 2020.
\bibitem{41}	L. Wen, L. Gao, and X. Li, “A new deep transfer learning based on sparse auto-encoder for fault diagnosis,” IEEE Trans. Syst. Man, Cybern. Syst., vol. 49, no. 1, pp. 136–144, Jan. 2019.
\bibitem{42}	M. Simon, E. Rodner, and J. Denzler, “ImageNet pre-trained models with batch normalization,” Dec. 2016.
\bibitem{43}	B. Neyshabur, H. Sedghi, and C. Zhang, “What is being transferred in transfer learning?,” Adv. Neural Inf. Process. Syst., vol. 2020-Decem, no. NeurIPS, 2020.
\bibitem{44}	T. Mensink, J. Uijlings, A. Kuznetsova, M. Gygli, and V. Ferrari, “Factors of Influence for Transfer Learning across Diverse Appearance Domains and Task Types,” 2021.
\bibitem{45}    M. Iman, J. A. Miller, K. Rasheed, R. M. Branch, H. R. Arabnia, “EXPANSE: A Deep Continual / Progressive Learning System for Deep Transfer Learning.” arXiv preprint arXiv:2205.10356, 2022.

\end{thebibliography}
\end{document}